# VISTA: Visualization of Token Attribution via Efficient Analysis

Syed Ahmed
syed_ahmed12@infosys.com

Bharathi Vokkaliga Ganesh
bharathi_ganesh01@infosys.com

Jagadish Babu P
jagadish.p02@infosys.com

Karthick Selvaraj
karthick.s50@infosys.com

Praneeth Talluri
praneeth.talluri@infosys.com

Sanket Hingne
sankethemraj.hingne@infosys.com

Anubhav Kumar
anubhav.kumar03@infosys.com

Anushka Yadav
anushka.yadav01@infosys.com

Pratham Kumar Verma
prathamkumar.verma@infosys.com

Kiranmayee Janardhan
kiranmayee.j@infosys.com

Mandanna A N
mandanna_an@infosys.com

Responsible AI Office

Infosys Limited, Bangalore, India

## ABSTRACT

Understanding how Large Language Models (LLMs) process information from prompts remains a significant challenge. To shed light on this "black box," attention visualization techniques have been developed [1], [14] to capture neuron-level perceptions and interpret how models focus on different parts of input data. However, many existing techniques are tailored to specific model architectures, particularly within the Transformer family, and often require backpropagation [5], resulting in nearly double the GPU memory usage and increased computational cost. A lightweight, model-agnostic approach for attention visualization remains lacking [25].

In this paper, we introduce a model-agnostic token importance visualization technique to better understand how generative AI systems perceive and prioritize information from input text, without incurring additional computational cost. Our method leverages perturbation-based [11] strategies combined with a three-matrix analytical framework to generate relevance maps that illustrate token-level contributions to model predictions. The framework comprises: (1) the Angular Deviation Matrix, which captures shifts in semantic direction; (2) the Magnitude Deviation Matrix, which measures changes in semantic intensity; and (3) the Dimensional Importance Matrix, which evaluates contributions across individual vector dimensions. By systematically removing each token and measuring the resulting impact across these three complementary dimensions, we derive a composite importance score that provides a nuanced and mathematically grounded measure of token significance. To support reproducibility and foster wider adoption, we provide open-source implementations of all proposed and utilized explainability techniques, with code and resources publicly available at https://github.com/Infosys/Infosys-Responsible-AI-Toolkit

## 1. INTRODUCTION

### 1.1 Background and Motivation

The emergence of Large Language Models (LLMs) including GPT [3], BERT [2], and their successive variants has fundamentally redefined the landscape of computational linguistics, enabling unprecedented performance across comprehension and text generation tasks. Despite their impressive performance across a wide array of tasks, the internal workings of these models remain largely opaque, often described as "black boxes [17]." This opacity creates a fundamental tension between model capability and practical deployability, as organizations in domains requiring auditability, such as patient care, financial regulation, and legal compliance.

Understanding how these models perceive and prioritize different parts of an input prompt is critical not only for transparency, but also for debugging, model auditing, and bias detection [15]. A key approach to demystifying model behavior has been attention visualization [22], which attempts to reveal how and where models focus during inference. While these techniques have offered valuable insights, they are often tightly coupled with specific architectures typically Transformers [14] and rely on attention weights [12] or backpropagation-based methods. These constraints introduce challenges: they demand deeper access to the model's internals, are often computationally intensive, and tend to double memory usage, limiting their practicality in real-time or resource-constrained applications [26].

To address these limitations, we propose a model-agnostic token importance visualization technique [6] that is lightweight, efficient, and architecture-independent. Our approach does not require any access to the model's internal gradients [8] or attention mechanisms. Instead, it leverages a perturbation-based strategy [11] paired with a multi-dimensional analytical framework to evaluate token importance. By systematically removing individual tokens from the input sentence and measuring the resulting change across three complementary matrices: Angular Deviation, Magnitude Deviation, and Dimensional Importance. This enables a measurable assessment of each token's contribution to the aggregate semantic representation of the input prompt, revealing which words exert the greatest influence on the embedding structure. The greater the shifts in the sentence's semantic representation across these dimensions, the more important the removed token is.

This technique is designed to be both computationally efficient and highly interpretable, rendering the approach practical for integration into production-grade generative AI pipelines without imposing significant computational overhead. Additionally, by filtering out common or semantically lightweight words, our method ensures that only the most impactful tokens are highlighted in the final relevance map.

1.2 Research Objectives

This paper details a methodology for calculating a token's importance score based on its impact on the prompt's collective semantic meaning. Our specific objectives are:

1. To develop a model-agnostic framework for token importance quantification
2. To capture multiple dimensions of semantic contribution (direction, intensity, and dimensional balance)
3. To provide mathematically rigorous and interpretable importance metrics
4. To establish a foundation for automated prompt optimization

1.3 Approach Overview

We utilize GloVe (Global Vectors for Word Representation) to transform tokens into a high-dimensional vector space. The core of our method is perturbation-based [11] analysis: we systematically ablate each token and quantify the resulting geometric and semantic shifts in the prompt's aggregate embedding through three complementary analytical matrices.

2. THEORETICAL FOUNDATION

2.1 Vector Space Semantics

Our methodology is grounded in the distributional hypothesis of linguistics, which posits that words appearing in similar contexts tend to have similar meanings. GloVe embeddings capture this principle by representing each word as a dense vector in a high-dimensional space, where geometric relationships (distances and angles) correspond to semantic relationships.

For a vocabulary $V$, GloVe learns a mapping:
$$w \in V \rightarrow E(w) \in \mathbb{R}^d$$

where $d$ is the embedding dimension (50 in our implementation).

## 2.2 Aggregate Prompt Representation

Given an input prompt $P$ consisting of $n$ tokens $T = \{t_1, t_2, \ldots, t_n\}$, we represent the entire prompt as a single aggregate vector. This representation is computed as the vector sum of individual token embeddings:

$$E_{orig} = \sum_{i=1}^{n} E(t_i)$$

This sum captures the collective semantic content of the prompt. While simple, this additive model has been shown to be effective in various NLP tasks and provides a tractable foundation for perturbation analysis.

## 2.3 Perturbation-Based Importance

Our methodology is founded on the principle that a token's importance is proportional to the semantic disruption its absence causes [11]. For each token $t_k$, we construct a "perturbed" prompt embedding by removing that token:

$$E_{pert_k} = E_{orig} - E(t_k) = \sum_{i=1}^{n} E(t_i) \; i \neq k$$

The difference between $E_{orig}$ and $E_{pert_k}$ quantifies the token's contribution to the prompt's overall semantic representation. However, a single scalar measure of this difference would be insufficient to capture the multifaceted nature of semantic contribution. Therefore, we decompose the impact into three distinct analytical dimensions.

## 3. METHODOLOGY: THREE-MATRIX ANALYTICAL FRAMEWORK

Our comprehensive token importance score is derived from three complementary matrices, each capturing a distinct aspect of how a token's removal affects the prompt's semantic representation.

## 3.1 THE ANGULAR DEVIATION MATRIX

### 3.1.1 Conceptual Foundation
The Angular Deviation Matrix measures how the removal of a token alters the overall semantic *direction* of the prompt in the embedding space. This component is critical because semantic direction encodes the primary topic or intent of the prompt. A significant change in direction implies the token was crucial for establishing the prompt's core meaning.

### 3.1.2 Geometric Interpretation
In vector space semantics, the angle between two vectors represents their semantic similarity. We quantify this directional relationship using the cosine function, which yields a scale-invariant similarity metric bounded between $-1$ $and$ $1$, unaffected by differences in vector length:

$$cos(\theta) = \frac{(v_1 \cdot v_2)}{(||v_1|| \times ||v_2||)}$$

This formulation isolates the angular separation between two vectors, capturing purely orientational semantic alignment.

When $cos(\theta) = 1$, the vectors point in identical directions (perfect semantic alignment). When $cos(\theta) = 0$, they are orthogonal (semantically unrelated). When $cos(\theta) = -1$, they point in opposite directions (antonymous or contradictory).

### 3.1.3 Mathematical Formulation
For token $t_k$, we calculate the cosine angle between the original prompt embedding $E_{orig}$ and the perturbed embedding $E_{pert_k}$:

$$cos(\theta_k) = \frac{(E_{orig} \cdot E_{pert_k})}{(||E_{orig}|| \times ||E_{pert_k}||)}$$

To transform this into an importance score where higher values indicate greater importance, we apply the following transformation:

$$Score_{angular_k} = \frac{(1 - cos(\theta_k))}{2}$$

This normalization ensures:
- $Score_{angular_k} \in [0, 1]$
- $Score_{angular_k} = 0$ when the token causes no directional change ($cos(\theta_k) = 1$)
- $Score_{angular_k} = 1$ when the token causes maximum directional change ($cos(\theta_k) = -1$)

3.1.4 Interpretation
A high angular deviation score indicates that removing the token causes a significant shift in the prompt's semantic orientation. Such tokens typically include:
- Primary subject nouns (e.g., "AI", "model", "system")
- Action verbs that define the task (e.g., "analyze", "generate", "classify")
- Domain-specific terminology that anchors the prompt's context

Conversely, low scores are assigned to tokens that do not substantially alter the direction, such as:
- Common stopwords (e.g., "the", "a", "an")
- Auxiliary verbs (e.g., "is", "are", "was")
- Generic modifiers that do not shift the core topic

3.1.5 Example Calculation
Consider the prompt: "The AI system processes natural language effectively"

For the token "AI":
1. $E_{orig}$ points in a direction representing the full semantic content
2. $E_{pert_{AI}}$ excludes "AI", shifting toward a more generic "system processes" meaning
3. The cosine angle between these vectors is significantly less than 1
4. The resulting $Score_{angular_{AI}}$ is high, reflecting "AI"'s importance in defining the prompt's topic

3.2 THE MAGNITUDE DEVIATION MATRIX

3.2.1 Conceptual Foundation

The Magnitude Deviation Matrix quantifies a token's contribution to the overall "semantic intensity" or "weight" of the prompt. In vector space models, the magnitude ($L2\ norm$) of an embedding vector can be interpreted as a measure of semantic salience or informativeness. Tokens that significantly increase or decrease this magnitude are considered important contributors to the prompt's overall semantic strength.

3.2.2 Geometric Interpretation
The $L2\ norm$ (Euclidean length) of a vector $v$ in $d$-dimensional space is defined as:

$$||v|| = \sqrt{\left(\sum_{i=1}^{d} v_i^2\right)}$$

In the context of word embeddings, a larger magnitude often correlates with more semantically rich or specific content. Generic or weak semantic content tends to result in lower magnitude aggregate vectors.

3.2.3 Mathematical Formulation
For token $t_k$, we calculate the absolute difference between the norms of the original and perturbed embeddings, normalized by the original norm:

$$Score_{magnitude_k} = \frac{|||E_{orig}|| - ||E_{pert_k}|||}{||E_{orig}||}$$

This normalization ensures:
- $Score_{magnitude_k} \in [0, 1]$
- The score is scale-invariant (independent of the absolute magnitudes)
- The score reflects the *relative* change in semantic intensity

3.2.4 Interpretation
A high magnitude deviation score indicates that the token substantially contributes to (or detracts from) the prompt's overall semantic weight. This can occur in two scenarios:

(a) Additive Contribution: The token's embedding vector is well-aligned with the rest of the prompt, amplifying the semantic signal.
   Example: In "very important task", the word "important" adds significant semantic weight to reinforce the urgency and priority.

(b) Cancellation Effect: The token's embedding vector opposes some components of the aggregate, reducing the net magnitude but potentially sharpening the semantic focus.
   Example: In "not insignificant", "not" reduces the magnitude of "insignificant" but refines the meaning.

Conversely, low scores indicate tokens whose removal has minimal effect on the overall semantic intensity:
- Tokens with small embedding magnitudes themselves
- Tokens whose embeddings are orthogonal to the main semantic direction
- Tokens that are redundant with other tokens in the prompt

### 3.2.5 Example Calculation
Consider the prompt: "The AI system processes natural language effectively"

For the token "effectively":
1. $||E_{orig}||$ represents the total semantic intensity of the prompt
2. $||E_{pert_{effectively}}||$ is slightly lower, as "effectively" adds semantic weight to the action "processes"
3. $Score_{magnitude_{effectively}} = \frac{||E_{orig}|| - ||E_{pert_{effectively}}||}{||E_{orig}||}$
4. This yields a moderate-to-high score, indicating "effectively" contributes meaningfully to the prompt's semantic strength

For the token "the":
1. $||E_{pert_{the}}|| \approx ||E_{orig}||$ (minimal change)
2. $Score\_magnitude\_the \approx 0$, reflecting "the"'s negligible contribution to semantic intensity

## 3.3 THE DIMENSIONAL IMPORTANCE MATRIX

### 3.3.1 Conceptual Foundation

While the Angular Deviation and Magnitude Deviation matrices analyze the aggregate vector as a whole, the Dimensional Importance Matrix examines the token's influence at the level of individual embedding dimensions. This fine-grained analysis captures the nuanced role of tokens in shaping the prompt's semantic profile across the entire 50-dimensional space.

Each dimension in the GloVe embedding space encodes a latent semantic feature (e.g., abstractness vs. concreteness, positive vs. negative sentiment, temporal vs. spatial reference). A token's importance can be understood not just by its overall contribution, but by *how* it contributes across these various semantic axes.

### 3.3.2 Dimensional Weighting Rationale
The key insight of this matrix is that a token's importance on a given dimension is context-dependent. A token is particularly important if:
- It contributes to a dimension where the prompt's aggregate vector has low energy (filling a semantic gap)
- It counteracts a dimension where the prompt has high energy (providing semantic balance or contrast)

This weighting strategy rewards tokens that either:
(a) Introduce novel semantic features not strongly present in the rest of the prompt
(b) Provide contrastive or qualifying information that balances the prompt's semantic profile

### 3.3.3 Mathematical Formulation
For each token $t_k$ and each dimension $j \in \{0, 1, \ldots, 49\}$, we define a weight function that captures the token's contribution to that dimension:

$$weight_j(t_k) = f(E_{orig}[j], E(t_k)[j])$$

The exact form of $f$ varies by implementation, but a common approach is:

$$weight_j(t_k) = |E(t_k)[j]| \times g(sign(E_{orig}[j]), sign(E(t_k)[j]))$$

where $g$ is a function that assigns higher weight when $E_{orig}[j]$ and $E(t_k)[j]$ have opposite signs (indicating the token provides contrast on this dimension).

The total dimensional importance score is the sum across all dimensions:

$$Score_{dimensional_k} = \sum_{j=0}^{49} weight_j(t_k)$$

3.3.4 Interpretation
A high dimensional importance score indicates that the token plays a crucial role in structuring the prompt's semantic representation across multiple latent
features. This often occurs for:
• Content words with rich, multi-faceted meanings (e.g., "analyze", "create",
  "evaluate")
• Domain-specific terms that activate unique semantic dimensions
• Modifiers that introduce important nuances (e.g., "not", "very", "partially")

Low scores typically correspond to:
• Tokens with sparse or generic semantic profiles
• Tokens that are redundant with other tokens across most dimensions
• Functional words that do not carry substantial semantic content

3.3.5 Dimensional Balance and Semantic Refinement
This matrix captures a phenomenon not addressed by the first two: the role of tokens in achieving semantic *balance*. Consider the prompt:
"The AI system does not process natural language effectively"

The token "not" may have:
• Low angular deviation (the prompt is still about AI processing language)
• Low magnitude deviation (negations often have relatively small magnitude)
• High dimensional importance (it flips the valence across multiple dimensions)

The Dimensional Importance Matrix ensures "not" receives appropriate weight despite low scores on the other two metrics.

3.3.6 Example Calculation
Consider the prompt: "The AI system processes natural language effectively"

For the token "processes":
1. Across the 50 dimensions, "processes" contributes significantly to action-oriented dimensions (verb semantics)
2. It may oppose or balance object-oriented dimensions (noun semantics) present in "AI", "system", "language"
3. The weighted sum across all dimensions yields a high $Score_{dimensional_{processes}}$
4. This reflects "processes" as a key action verb defining the prompt's task

For the token "the":
1. "the" has low values across most semantic dimensions (it's semantically weak)
2. It does not provide contrastive information on any dimension
3. $Score_{dimensional_{the}}$ is very low, confirming "the"'s minimal semantic contribution

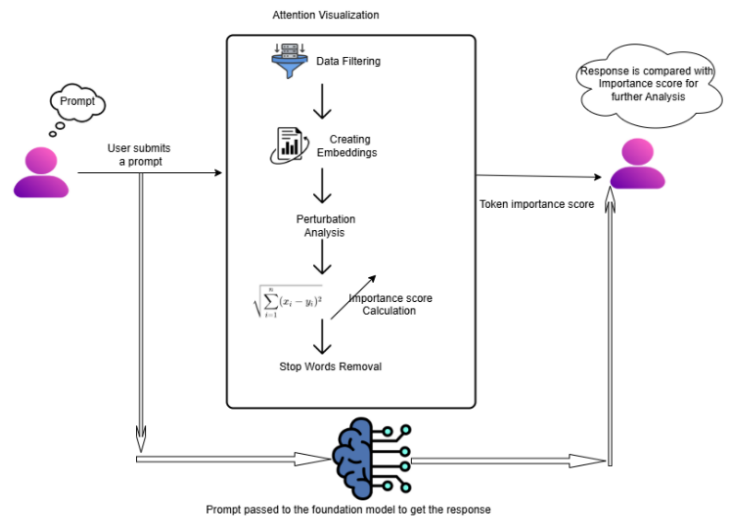

Figure 1: Token Importance

4. COMPOSITE IMPORTANCE SCORE

4.1 Integration Strategy

To derive a single, holistic measure of a token's importance, we combine the scores from the three analytical matrices. The choice of combination method is critical to ensure that the final score reflects a balanced assessment across all dimensions of importance. As stated in the image Figure 1 We employ multiplicative combination:

$$ImportanceScore_k = Score_{angular_k} \times Score_{magnitude_k} \times Score_{dimensional_k}$$

4.2 Rationale for Multiplicative Combination
Multiplicative combination offers several advantages:

(a) Robustness: A token must score well across *all three* dimensions to be considered highly important. Due to the multiplicative nature of this integration, a deficiency in any one dimension propagates through the entire composite, acting as a natural safeguard against inflated importance scores driven by a single metric in isolation.
(b) Semantic Completeness: True token importance requires contribution to direction (topic), intensity (weight), and dimensional balance (nuance). Multiplication captures this requirement for multi-dimensional significance.
(c) Dynamic Range: The multiplicative approach naturally creates a wide range of final scores, making it easier to distinguish between tokens of varying importance.
(d) Interpretability: If a token has a very low final score, one can examine which of the three component scores is the bottleneck, providing diagnostic insight.

4.3 Score Distribution and Interpretation
The final importance scores typically exhibit the following distribution:

• High Scores ($> 0.5$): Core content words that define the prompt's topic, task, and key entities. These are often:
  - Domain-specific nouns (e.g., "AI", "model", "data")
  - Primary verbs (e.g., "analyze", "generate", "predict")
  - Critical modifiers (e.g., "accurate", "comprehensive")

• Medium Scores ($0.2 - 0.5$): Supporting words that add important context or detail but are not central to the core meaning. These include:
  - Secondary nouns and verbs
  - Descriptive adjectives and adverbs
  - Connective words that structure the prompt

• Low Scores ($< 0.2$): Functional words with minimal semantic contribution:
  - Articles (e.g., "the", "a", "an")
  - Auxiliary verbs (e.g., "is", "are", "was")
  - Generic prepositions (e.g., "of", "in", "to")

4.4 Example: Complete Token Importance Analysis
Prompt: "The AI system processes natural language effectively"

Token Analysis:
1. "The"
  - $Score_{angular}$: 0.05 (minimal directional shift)
  - $Score_{magnitude}$: 0.03 (negligible intensity change)
  - $Score_{dimensional}$: 1.20 (low semantic content)
  - $ImportanceScore$: $0.05 \times 0.03 \times 1.20 = 0.0018$
  → *Very low importance* (functional word)

2. "AI"
  - $Score_{angular}$: 0.89 (major directional shift without "AI")
  - $Score_{magnitude}$: 0.76 (significant intensity contribution)
  - $Score_{dimensional}$: 15.67 (rich semantic profile)
  - $ImportanceScore$: $0.89 \times 0.76 \times 15.67 = 10.60$
  → *Very high importance* (core topic word)

3. "system"
  - $Score_{angular}$: 0.67 (moderate directional impact)
  - $Score_{magnitude}$: 0.54 (moderate intensity contribution)
  - $Score_{dimensional}$: 8.32 (moderate semantic richness)
  - $ImportanceScore$: $0.67 \times 0.54 \times 8.32 = 3.01$
  → *High importance* (key entity)

4. "processes"
  - $Score_{angular}$: 0.82 (substantial directional shift)
  - $Score_{magnitude}$: 0.68 (notable intensity contribution)
  - $Score_{dimensional}$: 12.45 (action verb with rich semantics)
  - $ImportanceScore$: $0.82 \times 0.68 \times 12.45 = 6.94$
  → *Very high importance* (core action)

5. "natural"
  - $Score_{angular}$: 0.45 (moderate directional impact)
  - $Score_{magnitude}$: 0.38 (moderate intensity contribution)

- $Score_{dimensional}$: 6.20 (qualifier with moderate semantics)
   - $ImportanceScore$: $0.45 \times 0.38 \times 6.20 = 1.06$
   → $Moderate\ importance$ (key qualifier)

6. "language"
   - $Score_{angular}$: 0.71 (significant directional shift)
   - $Score_{magnitude}$: 0.59 (notable intensity contribution)
   - $Score_{dimensional}$: 10.12 (domain-specific noun)
   - $ImportanceScore$: $0.71 \times 0.59 \times 10.12 = 4.24$
   → $High\ importance$ (core concept)

7. "effectively"
   - $Score_{angular}$: 0.52 (moderate directional impact)
   - $Score_{magnitude}$: 0.43 (moderate intensity contribution)
   - $Score_{dimensional}$: 7.80 (adverb with moderate semantics)
   - $ImportanceScore$: $0.52 \times 0.43 \times 7.80 = 1.74$
   → $Moderate\ importance$ (qualifier)

Final Ranking by Importance:
1. "AI" (10.60)
2. "processes" (6.94)
3. "language" (4.24)
4. "system" (3.01)
5. "effectively" (1.74)
6. "natural" (1.06)
7. "The" (0.0018)

5. GAM-BASED ENHANCEMENT FOR TOKEN IMPORTANCE

5.1 Motivation and Rationale While the three-matrix framework yields a composite importance score that is interpretable and efficient, it assumes a fixed multiplicative interaction between components. In practice, token contributions can exhibit non-linear relationships and context effects (e.g., position in the prompt) that are not fully captured by a single product. To model these effects without sacrificing interpretability, we incorporate a Generalized Additive Model (GAM). GAMs extend linear models by learning smooth functions over each feature and summing them, remaining transparent while capturing non-linearities.

5.2 Feature Set For each token $t$, we compute the following features (derived from Sections 3 and 4):

- $f_1$: Angular deviation $A(t) \in [0, 1]$
- $f_2$: Magnitude deviation $M(t) \in [0, 1]$
- $f_3$: Dimensional score $D(t) \geq 0$
- $f_4$: Position percentile $p(t) \in [0, 100]$ (token index normalized by prompt length)

5.3 Model Formulation The target is the percentile rank of the token within the importance sequence. The GAM is:

$$Percentile(t) = \beta_0 + s_1(A(t)) + s_2(M(t)) \\ + s_3(D(t)) + s_4(p(t)) + \varepsilon$$

where $s_1 \ldots s_4$ are smooth functions (e.g., cubic splines) learned from data, $\beta_0$ is the intercept, and $\varepsilon$ is the residual. This structure preserves additivity and interpretability: each term shows how a feature influences the percentile, independent of the others.

5.4 Training Data Generation

1. For each prompt $P$, compute importance scores via the current composite method (Section 4).
2. Sort tokens by the composite score and assign percentile ranks (100 = most important).
3. Record $(A(t), M(t), D(t), p(t), Percentile(t))$ for all tokens across diverse prompts/domains.

5.5 Inference Pipeline

Given a new prompt and its feature computations per token, the trained GAM outputs a predicted importance percentile for each token. Tokens are then ranked by predicted percentile.

5.6 Advantages of the GAM Enhancement
• Captures non-linear effects: importance may grow rapidly after thresholds in $A, M,$ or $D$.
• Position-aware: accounts for structural cues (lead vs. trailing tokens).
• Interpretable: Smooth functions can be plotted to show partial dependence per feature.
• Efficient: Inference is $O(k)$ per token ($k$ = number of features) and adds negligible overhead.

### 5.7 Example (Conceptual)

For the prompt "The AI system processes natural language effectively," tokens like 'AI' and 'processes' exhibit high $A, M,$ and $D$ and early positions, yielding $s_1 + s_2 + s_3 + s_4$ values near the upper range, hence high predicted percentiles. Functional words like 'The' show low $A$ and $M$ and low $D$, mapping to low percentiles.

### 5.8 Implementation Note

We use standard GAM tooling with spline terms and cross-validation; production code loads a serialized model and predicts percentiles within the API layer.

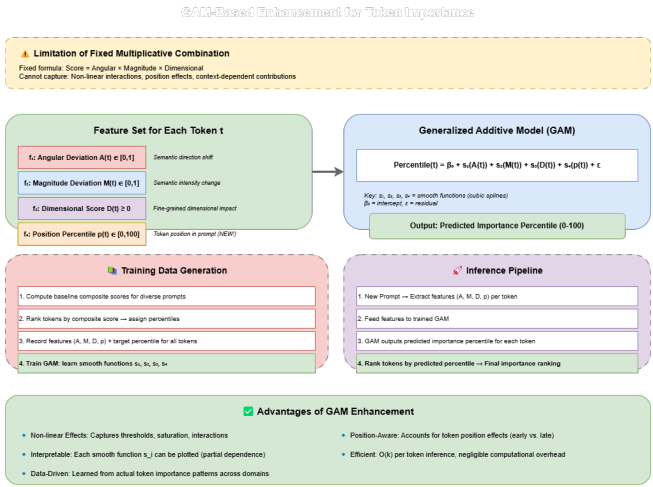

### 6. ALGORITHMIC IMPLEMENTATION

#### 6.1 Preprocessing Pipeline
Step 1: Tokenization
- Input prompt is split into tokens using whitespace and punctuation rules
- Each token is normalized (lowercase, stemming and removal of stopwords if required)

Step 2: Embedding Lookup
- Each token is mapped to its GloVe embedding vector
- Out-of-vocabulary tokens are handled via:
  (a) Character-level fallback
  (b) Subword tokenization
  (c) Zero vector (with appropriate logging)

Step 3: Aggregate Embedding Computation
- $E_{orig} = \Sigma\, E(t_i)$ for all tokens in the prompt

### 6.2 Core Computation Loop
For each token $t_k$ in the prompt:
 Step 1: Compute perturbed embedding
$$E_{pert_k} = E_{orig} - E(t_k)$$

 Step 2: Compute the directional similarity between the original and perturbed embeddings using the cosine function:
$$\cos \theta_k = \frac{\langle E_{orig}, E_{pert_k} \rangle}{(\|E_{orig}\|_2 \cdot \|E_{pert_k}\|_2)}$$
$$Score_{angular_k} = \frac{(1 - \cos_{theta_k})}{2}$$

 Step 3: Calculate Magnitude Deviation Score
$$Score_{magnitude_k} = \frac{|\,\|E_{orig}\| - \|E_{pert_k}\|\,|}{\|E_{orig}\|}$$

 Step 4: Calculate Dimensional Importance Score
  For each dimension $j$ in $[0, 49]$:
$$weight_{j_k} = compute_{dimensional_{weight}}(E_{orig}[j], E(t_k)[j])$$
$$Score_{dimensional_k} = \Sigma\, weight_{j_k}$$

 Step 5: Compute Composite Score
$$ImportanceScore_k = Score_{angular_k} \times Score_{magnitude_k} \times Score_{dimensional_k}$$

### 6.3 Computational Complexity
Time Complexity: $O(n \times d)$
 where $n$ = number of tokens, $d$ = embedding dimensionality (50)
 - Embedding lookup: $O(n)$
 - Per-token perturbation analysis: $O(n \times d)$
 - Total: Linear in both prompt length and embedding dimension

Space Complexity: $O(d)$
 - Storage for $E_{orig}$: $O(d)$
 - Storage for $E_{pert}$ (recomputed for each token): $O(d)$
 - Total: Constant w.r.t. prompt length

### 6.4 Optimization Strategies
(a) Vectorization: All matrix operations are implemented using NumPy for efficient SIMD execution

(b) Caching: GloVe embeddings are loaded once and cached in memory
(c) Parallel Processing: Token importance scores can be computed in parallel

## 7. SUMMARY AND GAP ANALYSIS: EXTENDING TOKEN IMPORTANCE TO QUALITY EVALUATION

While token importance analysis provides insights into individual token contributions within a single text, many real-world applications require comparing two texts to assess quality, completeness, and semantic alignment. One critical use case is automatic summarization, where a generated summary must faithfully represent the key information from a source document or context.

Conventional summary assessment methods, notably ROUGE, rely on surface-level lexical overlap through n-gram matching, which inherently overlooks deeper semantic correspondence between the source material and the generated summary. BLEU and METEOR consider surface-level similarity but are insensitive to missing critical concepts. BERTScore uses contextual embeddings but lacks interpretability and fine-grained diagnostic capabilities.

We extend our token importance methodology to address these limitations by introducing Summary and Gap Analysis, a framework that leverages token-level importance scores to:
1. Quantify semantic coverage of a summary relative to source content
2. Identify specific tokens or concepts that are missing or irrelevant
3. Provide actionable, interpretable feedback for summary improvement
4. Enable automated quality assessment without human annotation

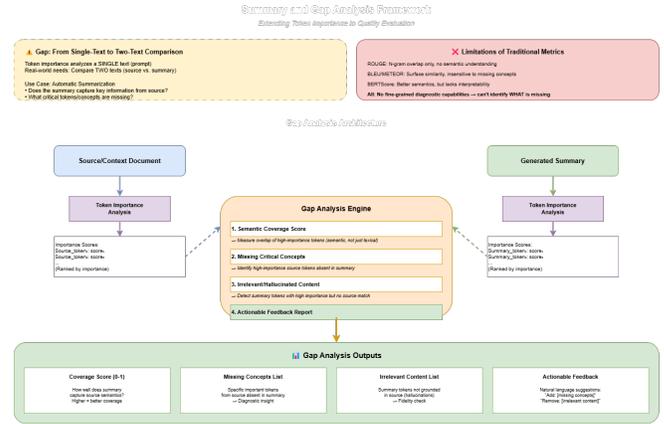

## 8. ADVANTAGES AND LIMITATIONS

### 8.1 Advantages
• Model-Agnostic: Does not require access to model internals or attention weights [6]
• Interpretable: Each component score has clear geometric and semantic meaning [17]
• Efficient: Linear time complexity enables real-time analysis
• Comprehensive: Captures multiple dimensions of token contribution
• Deterministic: Produces consistent results for the same input

### 8.2 Limitations
• Additive Assumption: Assumes prompt semantics can be modeled as vector sum
• Static Embeddings: GloVe does not capture context-dependent meaning [2]
• Token Independence: Does not explicitly model token interactions or dependencies [23]
• Language-Specific: Requires pre-trained embeddings for each language

## 9. CONCLUSION

This paper has presented a novel, multi-dimensional methodology for quantifying token importance in LLM prompts. By decomposing a token's contribution into three distinct analytical matrices, Angular Deviation, Magnitude Deviation, and Dimensional Importance, we provide a nuanced and mathematically rigorous framework for understanding prompt semantics.

The methodology offers a deterministic, model-agnostic, and interpretable [11] approach to token importance analysis. It serves as a valuable tool for:
- Researchers seeking to understand prompt structure and semantics
- Practitioners engineering effective prompts for LLM applications
- Organizations ensuring transparency and explainability in AI systems

The framework presented here represents a foundational step toward more transparent, controllable, and effective human-LLM interaction.